\documentclass[10pt]{article}
\usepackage{ijcnlp2011}
\usepackage[T1]{fontenc}
\usepackage{times}
\usepackage{url}
\usepackage{latexsym,amsmath,graphics,color,underlin}
\long\def\anonymize#1#2{#1}

\title{A Semantic Relatedness Measure Based on Combined Encyclopedic, Ontological and Collocational Knowledge}
\def\xxx{}
\author{\anonymize{\xxx Yannis Haralambous\\\xxx Institut Télécom -- Télécom Bretagne\\\xxx 
Département Informatique\\\xxx 
UMR CNRS 3192 Lab-STICC\\\xxx 
Technopôle Brest Iroise\\\xxx 
CS 83818, 29238 Brest Cedex 3, France\\\xxx 
\url{yannis.haralambous@telecom-bretagne.eu} \And\xxx  Vitaly Klyuev\\\xxx 
University of Aizu\\\xxx 
Aizu-Wakamatsu\\\xxx 
Fukushima-ken 965-8580, Japan\\\xxx 
\url{vkluev@u-aizu.ac.jp}}{Author names removed}}
\begin{document}
\maketitle              


\begin{abstract}
We describe a new semantic relatedness measure combining the Wikipedia-based Explicit Semantic Analysis measure, the WordNet path measure and the mixed collocation index. Our measure achieves the currently highest results on the WS-353 test: a Spearman $\rho$ coefficient of 0.79 (vs.\ 0.75 in \cite{gab2007}) when applying the measure directly, and a value of 0.87 (vs.\ 0.78 in \cite{Agirre})  when using the prediction of a polynomial SVM classifier trained on our measure.

In the appendix we discuss the adaptation of ESA to 2011 Wikipedia data, as well as various unsuccessful attempts to enhance ESA by filtering at word, sentence, and section level.
\end{abstract}

\section{Introduction}

\subsection{Semantic Relatedness and Corpora}

Semantic relatedness describes the degree to which concepts are associated via any kind of semantic relationship \cite{scriver}. Its evaluation is a fundamental NLP problem, with applications in word-sense disambiguation, text classification, information retrieval, automatic summarization and many other fields. In recent decades, a great variety of relatedness measures have been defined, based on corpora such as Wikipedia, Wiktionary, WordNet, etc.

Wikipedia is one of the most successful collaborative projects of all time. By a constantly growing number of additions, corrections and verifications, its contents grows in both quantity and quality, and is considered by many linguists as the corpus they had always dreamed of \cite{mining}.

By measuring the normalized tfidf values of words in a page, we can consider the page to be a weighted vector in the space of words. Inverting the matrix of these vectors we obtain weighted vectors of words in the space of pages. As every page deals with a single topic, we consider these vectors as being concept vectors. The ESA (Explicit Semantic Analysis) measure between two words is obtained by taking the cosine of their concept vectors \cite{gab2007}.

%
%

Unlike Wikipedia, WordNet \cite{wn}, a semi-formal lexical ontology \cite{otl}, has a fine and carefully-crafted ontological structure: word senses are represented by sets of synonyms (``synsets''), and there is a graph structure on synsets based on hypernymic relations. Several WordNet-based semantic relatedness measures have been defined, based on distances in the hypernymic graph, and often combined with word distribution in sense-tagged corpora. 

\subsection{Evaluation of Results, WS-353 Test}

\cite{ws353} introduce WS-353, a semantic relatedness test set consisting of 353 word pairs\footnote{Actually 352 pairs, since ``money~/ cash'' appears twice.} and a gold standard defined as the mean value of evaluations by up to 17 human judges. Although this test suite contains some quite controversial word pairs,\footnote{For example: ``Arafat~/ terror'' (0.765), ``Arafat~/ peace'' (0.673), ``Jerusalem~/ Israel'' (0.846), ``Jerusalem~/ Palestinian'' (0.765), etc.} it has been widely used in literature and has become the de facto standard for semantic relatedness measure evaluation.

Technically, the final result of the test is the Spearman $\rho$ rank correlation coefficient \cite{spear} between the relatedness ranking of pairs by human judges and that by the tested algorithm. So, in fact, it is not the value obtained for each pair that counts, but only the ranks.


\subsection{Our Approach}

By closely examining word pairs that failed to be ranked correctly by ESA, we came to the conclusion that the WS-353 word pairs belong (non-exclusively) to four classes, corresponding to different kinds of semantic relatedness and requiring different kinds of knowledge:


\begin{enumerate}
\item \emph{encyclopedic}: see Section~\ref{ency};


\item \emph{ontological}: see Section~\ref{onto};


\item \emph{collocational}: see Section~\ref{coll};


\item \emph{pragmatic}: see Section~\ref{pragma}.
\end{enumerate}



In this paper, we define a new semantic relatedness measure by combining knowledge related to these four classes.

\section{Encyclopedic Knowledge}\label{ency}

This class contains pairs that are best sorted by ESA. We note that \cite{Agirre} qualify ESA as a distributional approach. Indeed, technically two words are semantically related in ESA if they appear together frequently in Wikipedia pages. But since pages are descriptions of topics (= concepts), words are ESA-close when they appear frequently in common concept descriptions, and therefore in common semantic domains. Hence, ESA is semantically richer than a merely distributional approach.


ESA is the first and most important component of our combined relatedness measure. By adapting our implementation of the ESA algorithm to 2011 Wikipedia data (see App.~\ref{wp2011}), we obtain a Spearman $\rho=\text{0.7394}$. In the following sections we describe the components added to ESA in order to optimize its performance even further.

\section{Ontological knowledge}\label{onto}

\begin{figure}[ht]
\resizebox{\columnwidth}{!}{\includegraphics{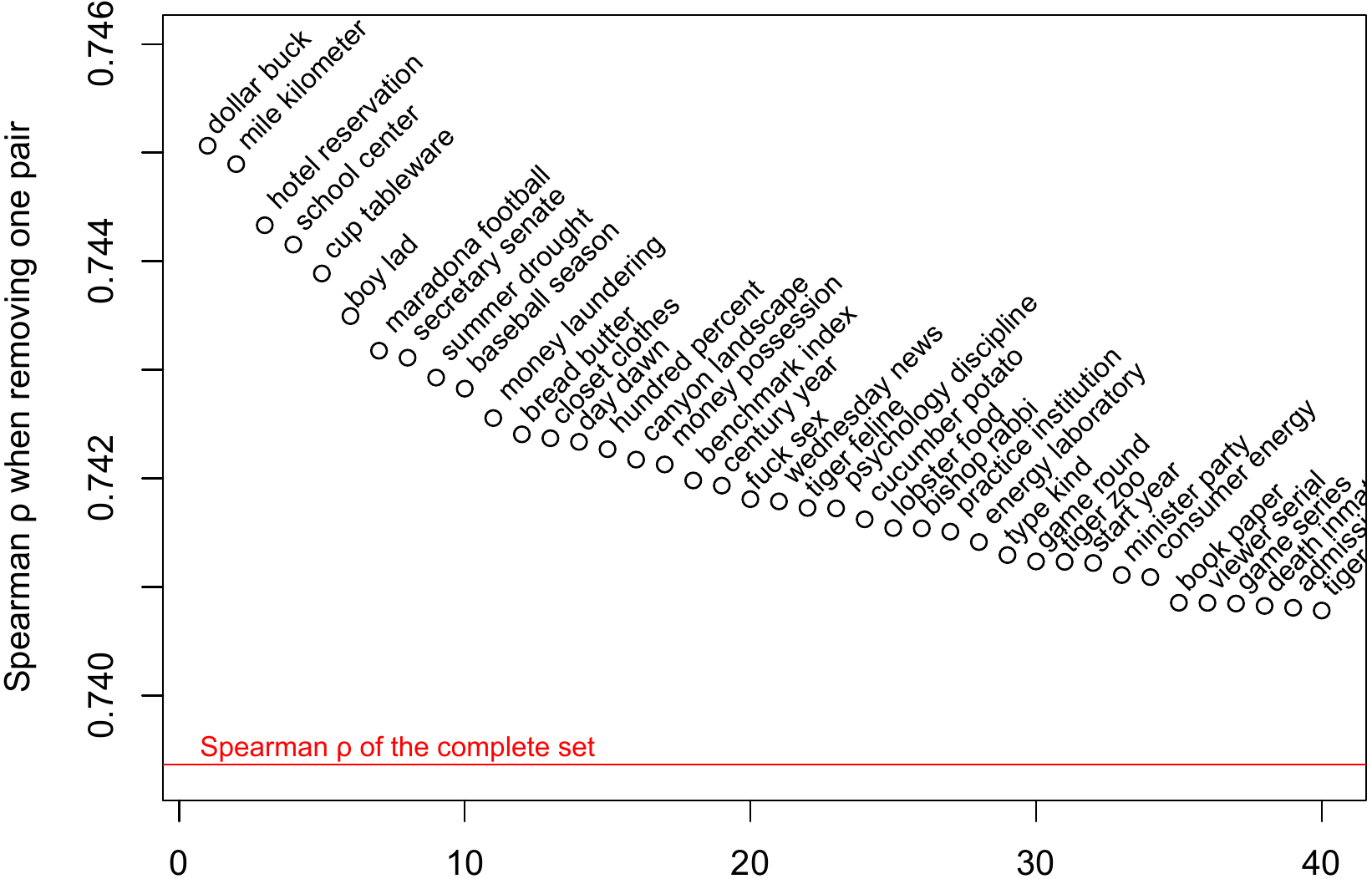}}
\caption{\fontsize{9.75pt}{10.75pt}\selectfont Spearman $\rho$ when removing a single pair from WS-353.\label{esa-1}}
\end{figure}

To get a better insight into the shortcomings of ESA on WS-353, we calculate Spearman $\rho$ for the WS-353 set minus a single pair, for every pair. In Fig.~\ref{esa-1} one can see the top 40 ``most problematic'' pairs: those whose removal increases $\rho$ the most. By taking a closer look at them we can get hints for further improvements of the measure.

First of all, we see pairs having a relation that is ontological in nature: ``tiger~/ feline'' (hyponym), ``mile~/ kilometer'' (coordinate terms, or ``classmates'' \cite{Torisawa2010vu}), ``dollar~/ buck'' (synonyms), etc. These relations are strong enough to justify the presence of the pairs in the test set, but do not necessarily imply high frequency of terms in common Wikipedia pages.

A good place for information of an ontological nature is WordNet. There have been several WordNet-based measures defined in the literature. When applying them\footnote{In fact, these measures apply to synsets rather than to words. To avoid going through a sense-disambiguation process, we take the optimistic approach of using for each pair of words, the pair of senses which are the most closely related. Hence, if $\hat{\mu}$ is a synset-measure, $s, s'$ are synsets and $w,w'$ words, we define the induced word-measure $\mu$ as $\mu(w,w'):=\max_{s\ni w,s'\ni w'}\hat{\mu}(s,s')$.} to the WS-353 test set we get the following~$\rho$:

{\small

\begin{center}
\begin{tabular}{|l|c|}\hline
WNP (Path-based) & 0.2873 \\
WUP \cite{wu} & 0.1356 \\
RES \cite{resnik} & 0.2112 \\
JCN \cite{jiang} & \textbf{0.3172} \\
LCH \cite{leacock} & 0.1437 \\
HSO \cite{hirst} & 0.1598 \\
LIN \cite{lin} & 0.1987 \\
LESK \cite{lesk} & 0.1304 \\\hline
\end{tabular}
\end{center}

}

Despite the fact that JCN (which combines WordNet-graph calculations and word frequencies from a corpus\footnote{For the distributional part of Jiang \& Conrath, Resnik and Lin, we use the Wikipedia 2011 corpus.}) rates best when used alone, the measure which we are going to use is WNP, which gives the best results when combined with ESA (see below). This measure is based exclusively on the shortest-path distance in WordNet and hence is purely ontological. For example, the WNP-measure of ``wood~/ forest'' is~1 (synonyms), ``bird~/ cock'' is 0.5 (hypernym), ``century~/ year'' is 0.33, ``bishop~/ rabbi'' is 0.25, etc.

We found that this measure provides bad results in its lower range (since the path length between distant nodes strongly depends on the density of WordNet for each knowledge domain). To understand the behavior of ESA and WNP measures in their low ranges, we progressively remove pairs from WS-353 \emph{in order of increasing relatedness}.

\begin{figure}[ht]
\resizebox{\columnwidth}{!}{\includegraphics{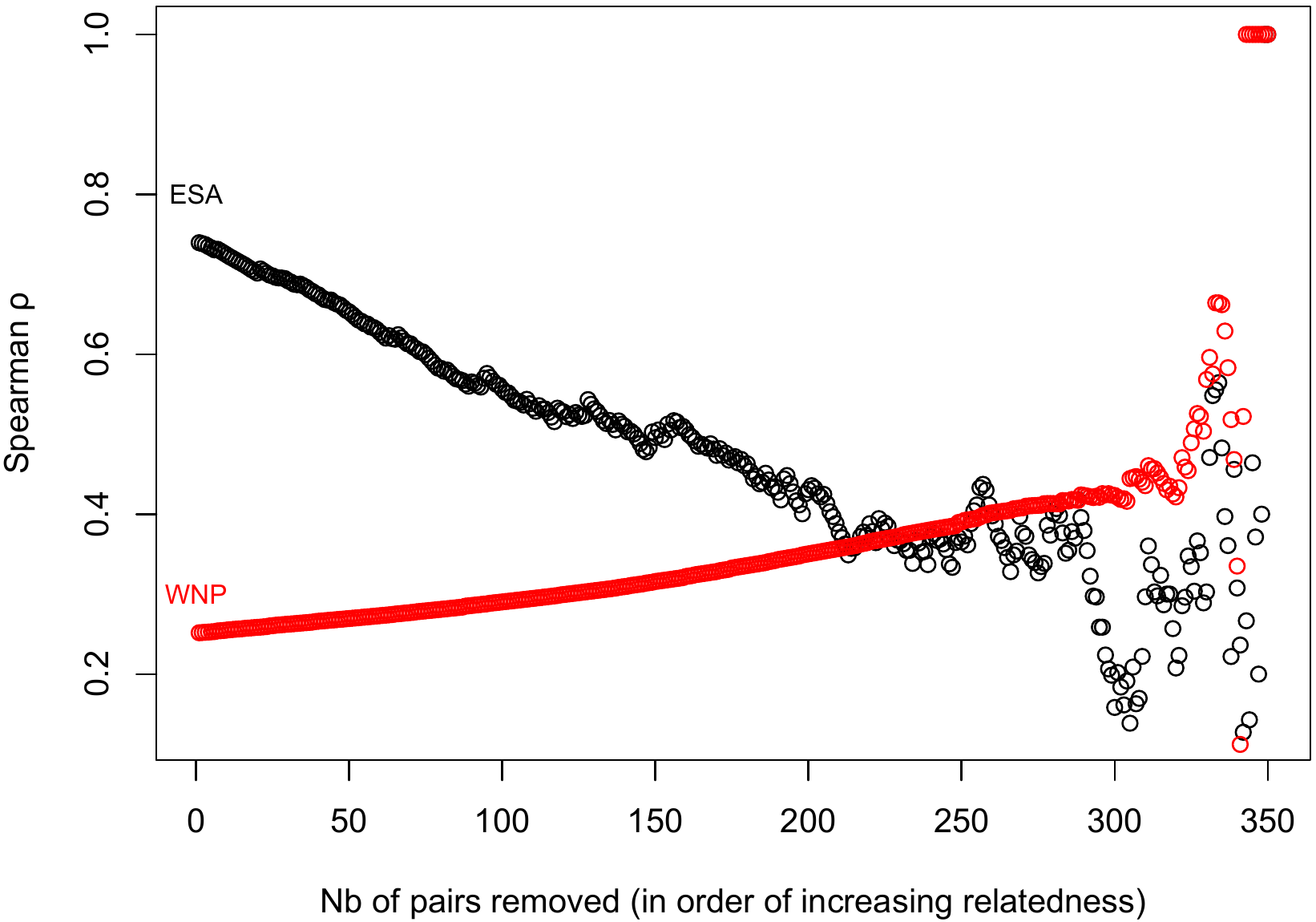}}
\caption{\fontsize{9.75pt}{10.75pt}\selectfont The effect on Spearman $\rho$ of the progressive removal of pairs in order of increasing relatedness, for ESA and WNP measures.\label{instab}}
\end{figure}

As we can see in Fig.~\ref{instab}, removing pairs in the small-value range of the measure \emph{strongly decreases} ESA (which, after half of the pairs are removed, becomes chaotic), while the same operation steadily \emph{increases} WNP. In other words, small-value pairs are crucial positive contributors for ESA, but rather negative contributors for WNP. For this reason, we use only the upper range of WNP, and ignore its results for low-valued pairs. To achieve a smooth ``fade-out'' of WNP's lower range we multiply it by a sigmoid logistic function. We hence define a new measure
\begin{equation}\label{eqEW}\begin{aligned}
\mu_{\mathrm{EW}}(w_1,w_2)&=\mu_{\mathrm{ESA}}(w_1,w_2)\\
&\quad\cdot(1+\lambda\sigma_{m,s}(\mu_{\mathrm{WNP}}(w_1,w_2))),
\end{aligned}\end{equation}
where $\lambda$ weights WNP with respect to ESA, $m$ is the sigmoid inflection point (=~a soft boundary of WNP's lower range), $s$ is the steepness of the sigmoid (small $s$ makes the central part of the sigmoid closer to a vertical line), and ``EW'' stands for ``ESA and WordNet.''

Calculations give the following optimal result:

{\small

\begin{center}
\begin{tabular}{|p{5cm}|l|}\hline
$\lambda=4.665,m=0.26,s=0.05$&$\rho=\text{\textbf{0.7779}}$\\
\hline
\end{tabular}
\end{center}

}

\noindent which surpasses the \cite{gab2007} ESA result of 0.75 by 5.2\%. The parameter values have been obtained by gradient descent. In the next section we will further enhance this result by taking collocations into account.

\section{Collocational Knowledge}\label{coll}

Returning to Fig.\ \ref{esa-1}, we see that many ``problematic'' pairs are in fact collocations: ``baseball~/ season,'' ``money~/ laundering'', ``hundred~/ percent,'' etc. We claim that the collocational nature of these word pairs has motivated their inclusion in WS-353. To show this, we calculated the collocation index (defined as $\frac{2\#(w_1w_2)}{\#(w_1)+\#(w_2)}$) of all WS-353 pairs\footnote{We obtained WS-353 pair and word frequencies from the 53.45 billion-word Google\-Books corpus \cite{googlebooks}. We considered only books published after 1970.}.

\begin{figure}[ht]
\resizebox{\columnwidth}{!}{\includegraphics{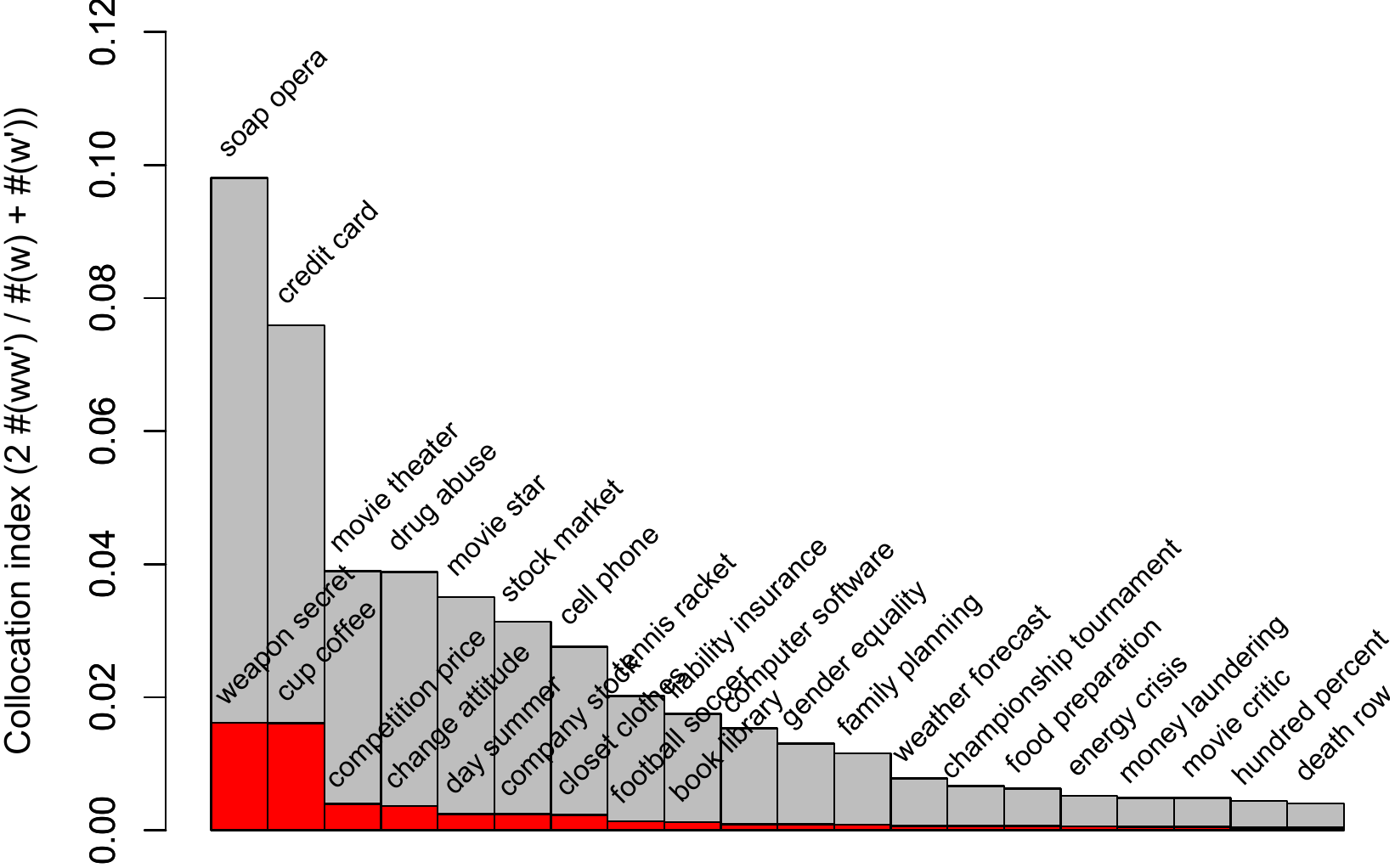}}
\caption{\fontsize{9.75pt}{10.75pt}\selectfont Top twenty direct (in gray) and inverse (in red) collocation indices for WS-353.\label{colloc}}
\end{figure}

The primary goal of WS-353 is to evaluate relatedness measures, and these are symmetric by definition (we always have $\mu(w_1,w_2)=\mu(w_2,w_1)$). If the word pairs were chosen on strictly semantic criteria, and if  collocations were purely accidental, then we would have a roughly equal number of pairs $(w_1,w_2)$ where $w_1w_2$ is a collocation and pairs where $w_2w_1$ is a collocation. 

Fig.~\ref{colloc} shows that this is not the case: for the word pairs concerned, WS-353 developers have almost systematically chosen to write the words in the order in which they form a collocation.

But neither ESA nor WNP recognize collocations: the former because of the bag-of-words principle underlying tfidf, and the latter only in the case where the collocational pair is a concept on its own. Indeed, most of the collocations in Fig.~\ref{colloc} are WordNet concepts (the exceptions being: ``gender~/ equality,'' ``food~/ preparation,'' ``secret~/ weapon,'' ``energy~/ crisis,'' etc.) but knowledge of that fact is not sufficient for ranking, since there is no mention in WordNet of the strength of the collocational relation.

We use the collocation index to further enhance our EW relatedness measure.

\begin{figure}[ht]
\resizebox{\columnwidth}{!}{\includegraphics{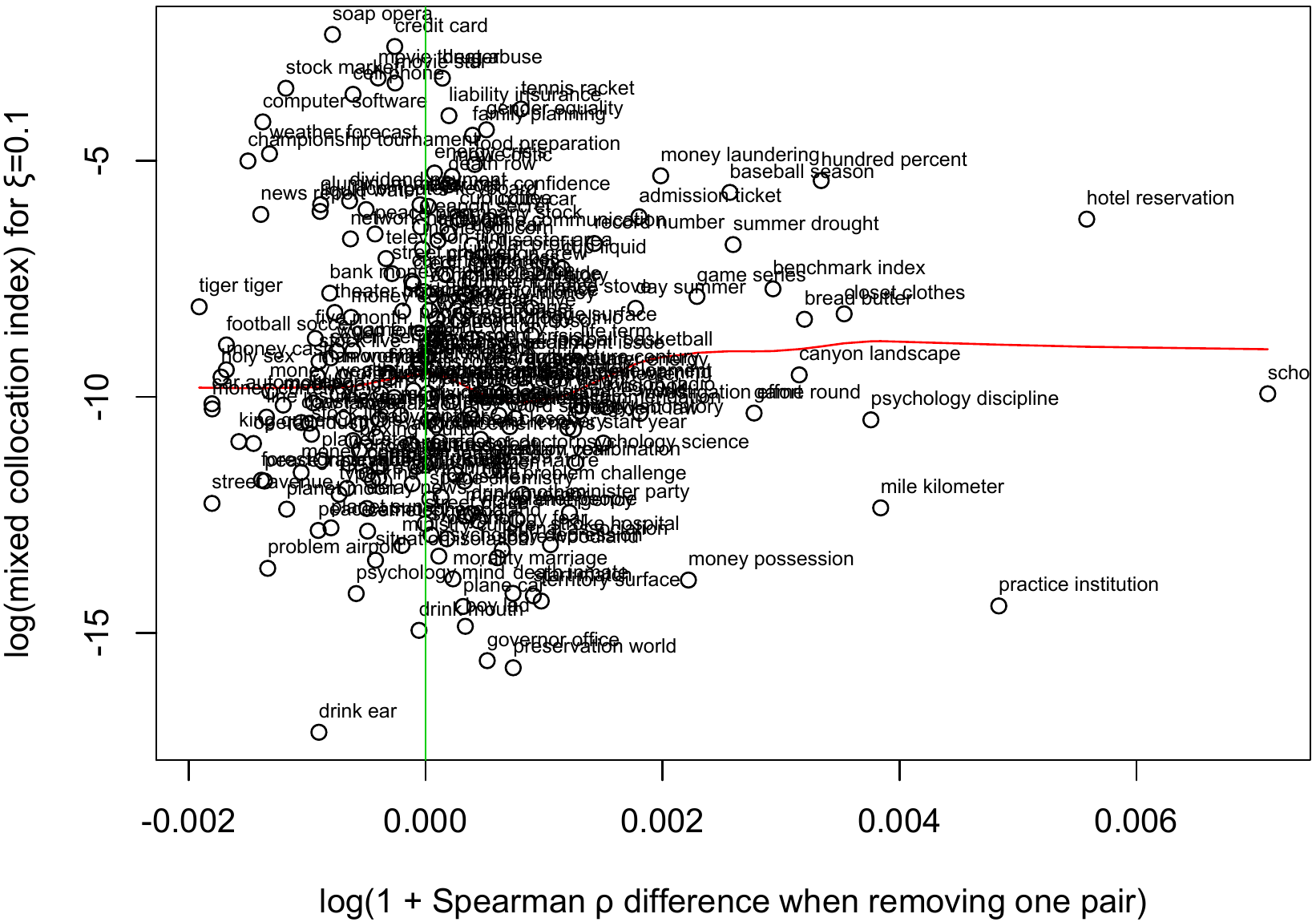}}
\caption{\fontsize{9.75pt}{10.75pt}\selectfont Collocation index vs. Spearman stability of EW. The red line is LOWESS polynomial regression \cite{Cleveland1981}.\label{colloc3}}
\end{figure}

Note that this index is \emph{not} a measure (for example, the collocation index of ``tiger~/ tiger'' is not~1) and cannot be used directly as such. 

How do collocational pairs contribute to the WS-353 Spearman $\rho$ value? In Fig.~\ref{colloc3} one can compare collocation index and Spearman stability (that is, the effect on~$\rho$ of the removal of a single word pair). Pairs located on the green vertical line are those whose removal does not affect Spearman $\rho$. Those on the right increase $\rho$ when removed. We observe that most collocations are on the right; in other words, they are negative contributors. The most problematic ones are collocations which are not individual WordNet concepts (typical examples: ``school~/ center,'' ``hotel~/ reservation,'' ``canyon~/ landscape,'' etc.). 

On the other hand, on the left side we find collocations that contribute positively to $\rho$: in many cases these have a strong ontological relation (``tiger~/ tiger,'' ``street~/ avenue,'' ``football~/ soccer,'' etc.) which is probably the main reason for their positive contribution. The LOWESS polynomial regression line is quasi-horizontal, so we cannot infer whether or not collocation index is correlated with~$\rho$.

\begin{figure}[ht]
\resizebox{\columnwidth}{!}{\includegraphics{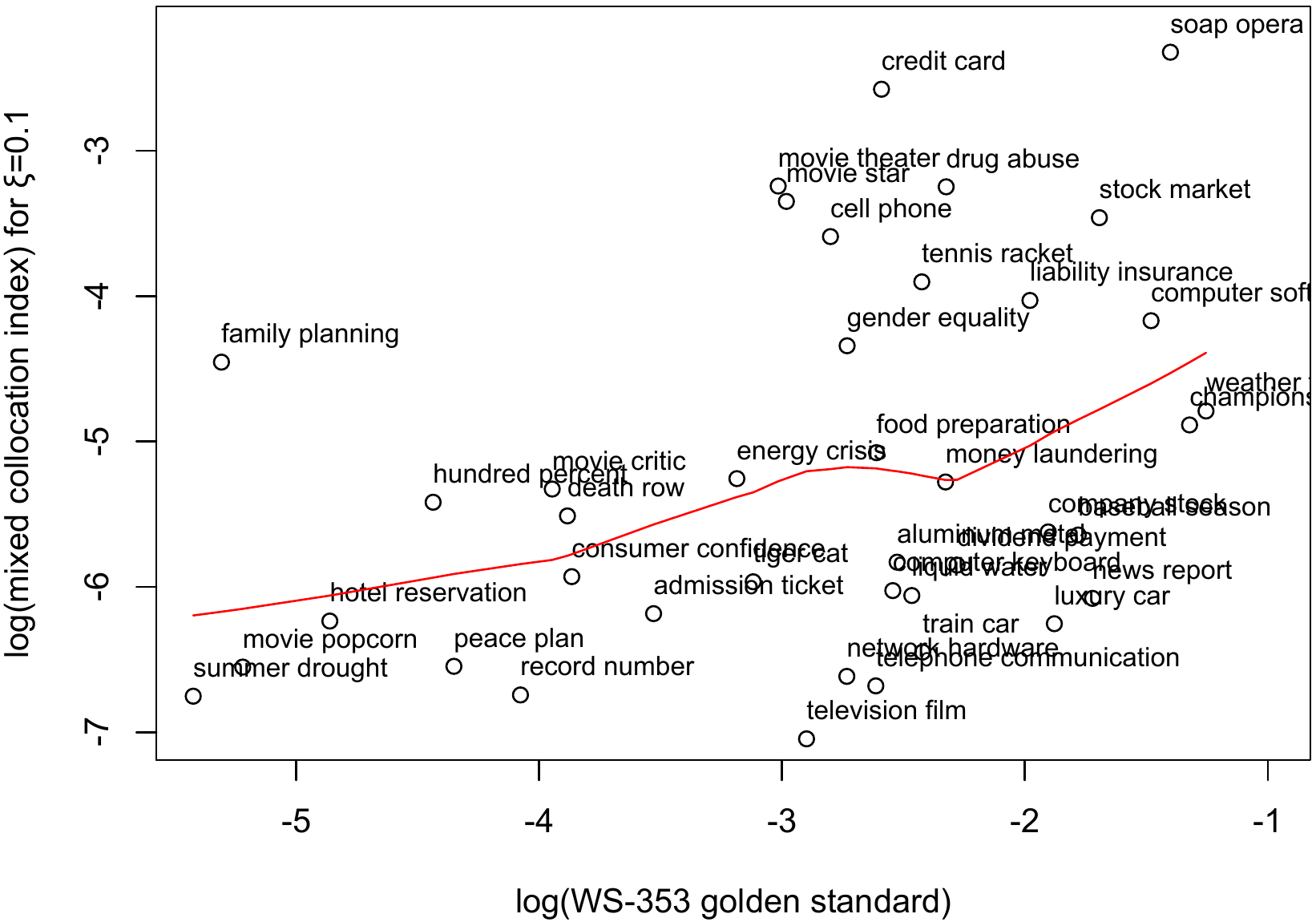}}
\caption{\fontsize{9.75pt}{10.75pt}\selectfont Collocation index vs. WS-353 gold standard. The red line is LOWESS polynomial regression.\label{colloc2}}
\end{figure}

An auxiliary question is whether collocation index values (at least in the high range) are correlated with the actual values of the WS-353 gold standard. 
Fig.~\ref{colloc2} compares these two quantities. As we can see, LOWESS polynomial regression is almost steadily monotonically increasing, which shows that, although not a measure per se, (high-range) collocation index could be useful for relatedness measurement. 

%

We combine the previously defined EW measure with collocation index, by defining measure EWC (=~``ESA + WordNet + collocations'') as follows:
\begin{equation*}\begin{aligned}
\mu_{\mathrm{EWC}}(w_1,w_2)&=\mu_{\mathrm{ESA}}(w_1,w_2)\\
&\quad\cdot(1+\lambda\sigma_{m,s}(\mu_{\mathrm{WNP}}(w_1,w_2)))\\
&\quad\cdot(1+\lambda'\sigma_{m',s'}(C_\xi(w_1,w_2))),
\end{aligned}\end{equation*}
where $\lambda,m,s$ are as in \eqref{eqEW}, $\lambda'$, $m'$ and $s'$ are similar, and the \emph{mixed collocation index} $C_\xi$ is defined as follows:
$$
C_\xi(w_1,w_2)=\frac{2\#(w_1w_2)}{\#(w_1)+\#(w_2)}+\xi\frac{2\#(w_2w_1)}{\#(w_1)+\#(w_2)}
$$
where $\#(.)$ is the frequency in the corpus.


Calculations give the following optimal result:
{\small

\begin{center}
\begin{tabular}{|p{5cm}|l|}\hline
${\lambda=5.16, \mu=0.25, \lambda'=48.7}$, ${\mu'=0.19, s=s'=0.05, \xi=0.55}$&$\rho=\text{\textbf{0.7874}}$\\\hline
\end{tabular}
\end{center}

}
\noindent which is 1.2\% higher than EW and, to the best of our knowledge, currently the highest result for WS-353 by a direct measure (not using a support vector machine). The parameter values have been obtained by gradient descent.

We can interpret this result as follows: \emph{the EWC measure works best when the lower fourth of WordNet measure and the lower fifth of collocation index values are ignored, and when inverse collocations count half as much as direct ones}.

\section{Supervised Approach Using an SVM}

\cite[p.~25]{Agirre} train an SVM on pairs of WS-353 pairs; this allows them to get an insight on performance increase obtained by combining various measures. By combining knowledge from a Web corpus and from WordNet, they obtain a highest value of Spearman $\rho=\text{0.78}$. We calculated predictions of (4th degree polynomial) SVMs based on our EW and EWC measures, and obtained the following results, using 10-fold cross-validation:

{\small
\begin{center}
\begin{tabular}{|l|l|l|}\hline
Measure&Result\\\hline
EW (ESA + WNP)&$\rho=\text{0.7996}$\\
EWC (ESA + WNP + collocations)&$\rho=\text{\textbf{0.8654}}$\\\hline
\end{tabular}
\end{center}

}

We observe that even without collocations we already get a better value than \cite{Agirre}, and also that the collocation component increases this value significantly, hence validating our choice of using collocational knowledge to enhance semantic relatedness measurement.

\section{Pragmatic Knowledge}\label{pragma}

This class contains pairs not captured by the previous methods. The typical example is ``hotel~/ reservation'': its ESA value is very low, there is no ontological relation, and the collocation index is quite low as well. To capture the relatedness of such a pair, we need specific knowledge domain ontologies, providing relations such as ``A is part of a functional process of system~B'' (in this case: ``a `reservation' is part of the process of renting a room in a `hotel'\,''). We leave this as an open task for future development.

\section{Conclusion}

By combining two pre-existing semantic relatedness measures and by adding a component based on frequency of collocations, we have obtained a new measure that surpasses the one given in \cite{Agirre} by 11\% (when comparing results obtained by SVMs). We conjecture that this measure can further be enhanced by using pragmatic knowledge taken, for example, from specialized domain ontologies.

\fontsize{10pt}{11pt}\selectfont

\section*{Appendices}\appendix

\section{Adapting ESA to 2011 Wikipedia}\label{wp2011}

The original (and unreleased) C++ ESA implementation \cite{gab2007} is based on 2005 Wikipedia data (2.2\,GB) and achieves a Spearman $\rho=0.75$. A later implementation in Python and Java \cite{calli}, based on the same corpus, achieves $\rho=0.74$. We implemented ESA in Perl and similarly obtained $\rho=0.7404$ when based on 2005 data. The same algorithms applied to 2011 data (31\,GB), produced a disappointing $\rho=0.7047$. Indeed, between 2005 and 2011, Wikipedia has evolved as follows:

{

\fontsize{8pt}{8pt}\selectfont

\begin{center}
\begin{tabular}{|l|c|c|}\hline
&2005&2011\\\hline
\#concepts&866,881&4,178,454\\
\#terms/concept&96.1971&97.4243\\\hline
\end{tabular}
\end{center}

}

\noindent where by ``concepts'' we mean Wikipedia pages in the main namespace, and by ``terms,'' distinct stemmed words.

\begin{figure}[ht]
\resizebox{\columnwidth}{!}{\includegraphics{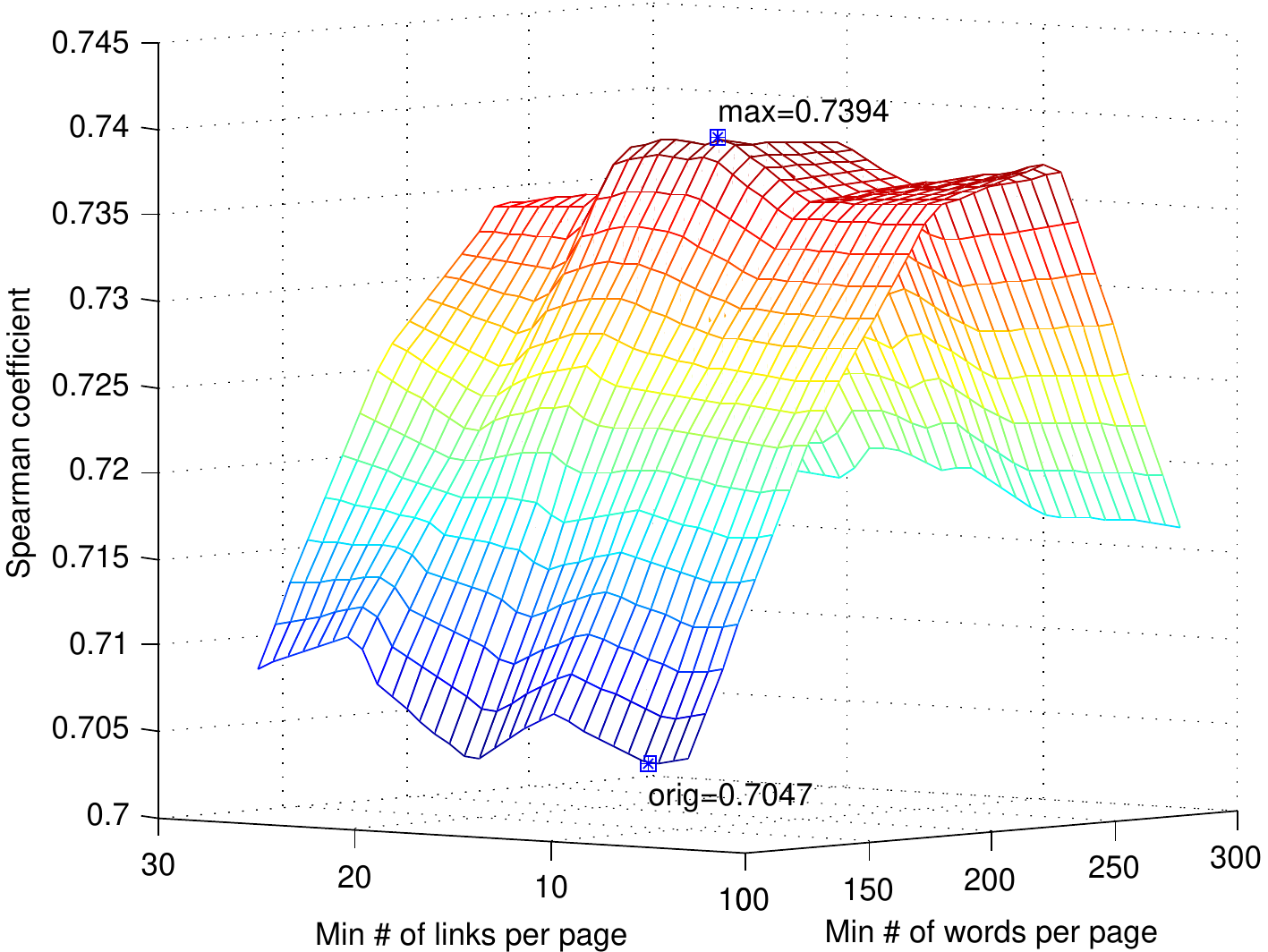}}
\caption{\small Adapting ESA to 2011 Wikipedia data by increasing the minimum number of distinct (stemmed) words and of in- and outgoing links per page.\label{matlab}}
\end{figure}

Following advice by Gabrilovich (personal communication), we increased the generality of concepts by filtering Wikipedia pages by two criteria: minimum number of terms, and minimum number of in- and outgoing links. The original values were: 100 terms and 5 links; by requiring a minimum of 200 terms and 14 links, we have attained the 2005 $\rho$ value (more precisely: $\rho=\text{\textbf{0.7394}}$). Fig.~\ref{matlab} displays $\rho$ as a function of our two criteria. 

In the following table, the column 2011 displays the results with original ESA setting, 2011* the ones with modified settings, $\overline{\text{df}}$ is mean document frequency of terms and term density is $\frac{\overline{\text{df}}}{\text{\#concepts}}$:

{

\fontsize{8pt}{8pt}\selectfont

\begin{center}
\begin{tabular}{|l|c|c|c|}\hline
&2005&2011&2011*\\\hline
\#concepts&132,689&311,209&155,767\\
\#terms/concept&165&279&414\\
\#terms&187,971&503,368&408,299\\
$\overline{\text{df}}$&116.3307&173.7199&159.0395\\
term density&0.00088&0.00056&0.00102\\\hline
\end{tabular}
\end{center}

}

As we see, terms are less densely distributed in the 2011 corpus, since the increase of their mean document frequency, though important, is overruled by an even more important increase in the number of concepts. By more efficiently pruning concepts and leaving $\overline{\text{df}}$ relatively stable, we manage to increase term density anew and hence, enhance performance.

\section{Experiments}

\cite{giraud} emphasize the importance of sharing negative results. Responding to their call, here are some of our failed attempts at increasing ESA performance on the 2005 corpus. Note that the standard ESA value we challenge is $\rho=\text{0.7404}$.

\subsection{At the Word Level: Lemmatization and POS Filtering}
ESA removes stop words and words with fewer than three letters before applying the Porter stemmer thrice. Instead of stemming, we lemmatized and then applied two strategies: keeping only nouns and proper names (Penn tags NN, NNP, and plurals), or also verbs and adjectives (tags starting with NN, NNP, VB, and JJ). Here are the results obtained:

{\small
\begin{center}
\begin{tabular}{|l|l|}\hline
Penn tags NN, NNS, NNP, NNPS&$\rho=\text{0.7194}$\\
Penn tags NN*, NNP*, VB*, JJ*&$\rho=\text{0.7178}$\\\hline
\end{tabular}
\end{center}
}
\noindent The performance loss is due to lemmatization, proving once again that while Porter stemming may seem a brutal technique, it works better than anything else. Note that, surprisingly, when adding verbs and adjectives we get a (slightly) \emph{smaller}~$\rho$.

\subsection{Filtering at the Sentence Level}
We attempted to triple the weight of sentences containing either the page title, or one of the (non stop-)words of the page title, or one of the anchors pointing to the page. This operation affected 1,399,165 sentences. Here are the results obtained:

{\small
\begin{center}
\begin{tabular}{|l|l|}\hline
Tripling weight of selected sentences&$\rho=\text{0.7293}$\\\hline
\end{tabular}
\end{center}
}

\subsection{Filtering at the Section Level}
%
The idea is to avoid ``historical sections'' in pages describing current notions or objects. Historical sections are detected by a higher frequency of past-tense verbs, unless of course the whole page is of a historical nature, and hence using primarily the past tense. Let $\pi=\frac{\text{\# past-tense verbs}}{\text{\# verbs}}$ for each Wikipedia page. We pruned sections of $\pi\geq0.8$ when the page had $\pi<0.8$. We also pruned sections named ``History,'' ``External links,'' ``References,'' ``See also,'' ``Further reading,'' and ``Bibliography.'' This affected 111,028 sections out of 470,948. Here are the results obtained:

{\small
\begin{center}
\begin{tabular}{|l|l|}\hline
Pruning of ``historical'' and other sections&$\rho=\text{0.6608}$\\\hline
\end{tabular}
\end{center}
}

\section{Implementation Details}

Implementation of ESA was done from scratch in Lex and Perl. To access WordNet v3, we used the Perl module \texttt{WordNet::Similarity} \cite{ws}. SVM calculations as well as 2D figures were done in R, and the 3D figure in Matlab. For lemmatizing and POS-tagging, we used TreeTagger \cite{treetagger}. Our code is publicly available at \anonymize{\url{http://omega2.enstb.org/yannis/similarity.php}}{URL removed}.

\medskip

\anonymize{\section{Acknowledgments}

We wish to thank Evgeniy Gabrilovich and Ça\u{g}atay Çall\i\ for their help in implementing ESA, and Sophia Ananiadou for her helpful advice.}{Acknowledgements section removed}

\fontsize{9.5pt}{10.5pt}\selectfont

\nocite{*}
\bibliographystyle{acl}
\bibliography{aaa}

\end{document}